\documentclass[twoside,leqno,twocolumn]{article}

\usepackage[letterpaper]{geometry}

\usepackage{ltexpprt}
\usepackage{hyperref}

\usepackage[square,numbers]{natbib}
\usepackage{amsmath}
\usepackage{soul}
\usepackage{xcolor}
\usepackage[export]{adjustbox}
\usepackage{listings}
\usepackage{bm}
\usepackage{array,multirow,graphicx}
\usepackage{diagbox}
\usepackage[
  separate-uncertainty = true,
  multi-part-units = repeat
]{siunitx}
\usepackage{enumitem}
\usepackage{makecell}
\usepackage{tabularx}

\usepackage{mathtools}
\usepackage[T1]{fontenc}
\usepackage{algorithm}
\usepackage{algorithmic}

\usepackage{lipsum}
\let\OLDthebibliography\thebibliography
\renewcommand\thebibliography[1]{
  \OLDthebibliography{#1}
  \setlength{\parskip}{0pt}
  \setlength{\itemsep}{0pt plus 0.3ex}
}

\begin{document}

\newcommand\relatedversion{}

\title{\Large Hierarchical Conditional Multi-Task Learning for Streamflow Modeling \relatedversion}

\author{Shaoming Xu \thanks{University of Minnesota.  \{xu000114,renga016,khand035, ghosh128,lixx5000,lichengl,tayal,jinzn,nieber,kumar001\}@umn.edu}
\and Arvind Renganathan \footnotemark[1]
\and Ankush Khandelwal \footnotemark[1]
\and Rahul Ghosh \footnotemark[1]
\and Xiang Li \footnotemark[1]
\and Licheng Liu \footnotemark[1]
\and Kshitij Tayal \footnotemark[1]
\and Peter Harrington \thanks{Lawrence Berkeley National Laboratory. pharrington@lbl.gov}
\and Xiaowei Jia \thanks{University of Pittsburgh. xiaowei@pitt.edu}
\and Zhenong Jin \footnotemark[1]
\and Jonh Nieber \footnotemark[1]
\and Vipin Kumar \footnotemark[1]
}

\date{}
\maketitle




\fancyfoot[R]{\scriptsize{Copyright \textcopyright\ 2025\\
Copyright for this paper is retained by authors}}



\begin{abstract}
Streamflow, vital for water resource management, is governed by complex hydrological systems involving intermediate processes driven by meteorological forces. While deep learning models have achieved state-of-the-art results of streamflow prediction, their end-to-end single-task learning approach often fails to capture the causal relationships within these systems. To address this, we propose Hierarchical Conditional Multi-Task Learning (HCMTL), a hierarchical approach that jointly models soil water and snowpack processes based on their causal connections to streamflow. HCMTL utilizes task embeddings to connect network modules, enhancing flexibility and expressiveness while capturing unobserved processes beyond soil water and snowpack. It also incorporates the Conditional Mini-Batch strategy to improve long time series modeling. We compare HCMTL with five baselines on a global dataset. HCMTL's superior performance across hundreds of drainage basins over extended periods shows that integrating domain-specific causal knowledge into deep learning enhances both prediction accuracy and interpretability. This is essential for advancing our understanding of complex hydrological systems and supporting efficient water resource management to mitigate natural disasters like droughts and floods. 
\end{abstract}
\section{Introduction} \label{sec:intro}
Streamflow modeling, which predicts the volume of water flowing through a river's cross-section over time, plays a key role in water resource management. These models have been widely used for managing reservoirs, hydropower operations, and flood control systems, as well as for guiding emergency responses during flood events. Additionally, streamflow models are important for understanding short-term and long-term climate change impacts, where daily or monthly forecast can aid in developing adaptive water management strategies.

Despite its importance, modeling streamflow in the Earth system is challenging, as its generation involves a complex interplay between meteorological factors (e.g., rainfall and temperature) and geophysical characteristics (e.g., soil, terrain, and land cover). This interplay is particularly important for understanding and modeling flow dynamics in a basin, the basic unit of hydrological study. Traditional process-based (PB) models \cite{bieger2017introduction, saad2019estimates} simulate streamflow based on physical laws and hydrological knowledge and have long been a cornerstone in hydrology. 
While PB models can be highly accurate when calibrated to a specific basin, their effectiveness decreases when applied to new basins due to basin uniqueness where extensive recalibration and adjustments are needed. Normally, the calibration process is expertise-driven and computationally intensive, limiting the scalability of traditional models for large-scale or multi-region applications.

The advancement of large-sample hydrology \cite{addor2020large} has provided publicly available datasets spanning hundreds to thousands of basins worldwide. 
Machine learning (ML) models trained on data from multiple basins can capture rainfall–runoff behaviors across diverse catchments. The hydrologic processes in each basin complement one another, enabling the model to improve predictions for individual basins by leveraging the collective information from all basins.
In recent studies, Recurrent Neural Networks (RNNs), particularly Long Short-Term Memory (LSTM) networks, excel at modeling dynamic systems influenced by past states and present inputs, making them ideal for streamflow prediction. Studies \cite{kratzert2019towards, kratzert2019ungauged} have shown that an LSTM trained on 531 basins in the United States significantly outperforms PB models. While tree-based models perform similarly on small datasets, LSTM models excel with larger datasets, where their ability to capture complex temporal dependencies becomes more advantageous \cite{gauch2021rainfall}. LSTM models have also been deployed in operational settings, such as Google's Flood Hub, providing reliable flood forecasts with up to five days of lead time \cite{nearing2024global}.

\begin{figure*}[htbp]
\centering
\includegraphics[width=\linewidth]{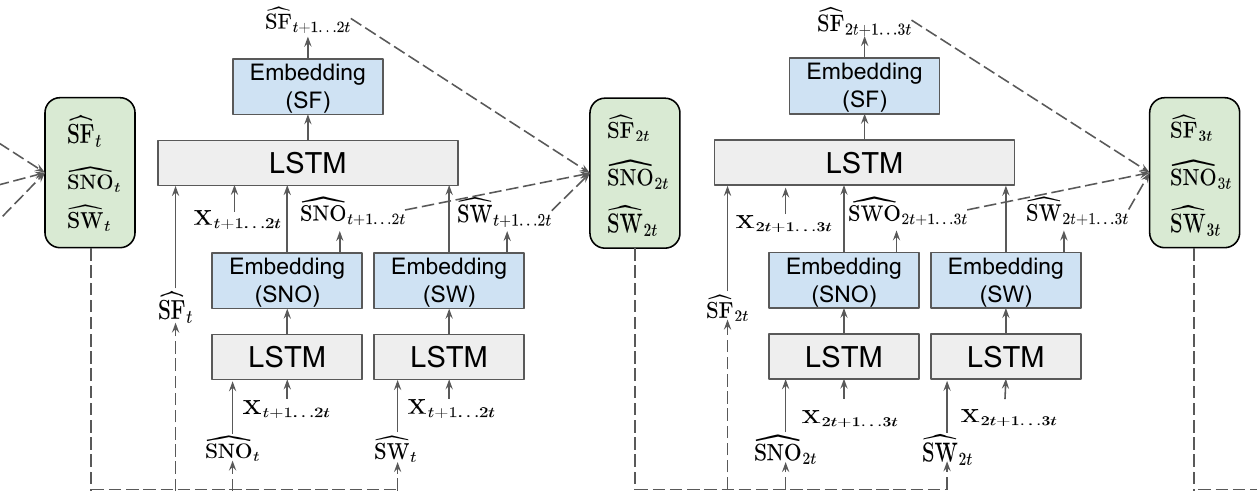}
\vspace{-1\baselineskip}
\caption{Hierarchical Conditional Multi-Task Learning for streamflow (SF) modeling during inference. Each segment contains t time steps, with the predicted Snowpack (SNO) and Soilwater (SW) from the current segment used to initialize the next.}
\label{fig:hcmtl}
\vspace{-1\baselineskip}
\end{figure*}

Although ML models have achieved state-of-the-art performance in streamflow prediction, they are typically implemented using single-task learning (STL), which does not explicitly incorporate domain knowledge regarding hydrological processes. These processes involve key intermediate variables, such as soil water and snowpack, which significantly influence how streamflow responds to meteorological drivers. However, STL ignores these intermediate variables and focuses solely on predicting streamflow, preventing ML models from capturing the underlying hydrological processes. This can reduce prediction accuracy, interpretability, and limits the model's ability to generalize across diverse basins. Additionally, streamflow data are long time series that often need to be divided into shorter segments to reduce computational complexity and mitigate vanishing/exploding gradient issues in model training \cite{bengio1993problem,pascanu2013difficulty}. However, existing models typically treat these segments independently, failing to capture their interactions and temporal dependencies, leading to suboptimal performance.

To address these challenges, we propose Hierarchical Conditional Multi-Task Learning (HCMTL) to manage multiple related tasks by leveraging their shared inter-dependencies. HCMTL creates separate network modules for each task, connecting them based on domain-specific causal relations. The hierarchical architecture reduces interference by separating unrelated tasks and enhances synergy by grouping related ones. It also allocates smaller networks for simpler tasks and larger networks for more complex ones, improving prediction reliability and acceptance by domain experts. Instead of using task outputs to connect the task modules, HCMTL employs task embeddings that capture the multi-dimensional internal states of intermediate tasks. These embeddings are more adaptable to unobserved processes, as they can encode broader patterns beyond explicitly modeled task outputs. Additionally, HCMTL incorporates Conditional Mini-Batch (CMB) algorithms \cite{xu2023mini}, which retain interactions between adjacent time series segments and preserve long-term dependencies in daily time series spanning years. These design choices improve HCMTL's ability to model real-world temporal dynamics with greater accuracy.

In this study, we apply the HCMTL to model the streamflow using the basin and climate attributes, meteorological variables, and two key intermediate target variables including snowpack and soil water. The selection of these two intermediate variables is guided by hydrological knowledge. Streamflow in a basin is regulated by its hydrological memory state, which can be snowpack during frozen seasons or soil water during regular periods, depending on the season of year. 
We verify the effectiveness of HCMTL on three regions including United States, Great Britain, and Central Europe from the CARAVAN~\cite{kratzert2022caravan}, which is a publicly available real-world hydrology benchmark dataset for streamflow modeling. 
We have released our codes and datasets needed to reproduce our results \footnote{\url{https://drive.google.com/drive/folders/1W0JEC0Ov3bu-FC2HxAtBOXrsTtDiNLc3?usp=drive_link}}. 
\section{Hierarchical Conditional Multi-Task Learning (HCMTL)} \label{sec:HCMTL}
We propose HCMTL to map meteorological variables and basin static characteristics to streamflow observations. HCMTL consists of three key components: a hierarchical multi-task network, task modules connected through network embeddings, and conditional mini-batch learning. These features make HCMTL a robust framework for accurately modeling streamflow across diverse regions over extended time periods.

\subsection{Hierarchical Multi-Task Network (HMTnet)} 
Natural systems are often complex, involving multiple intermediate processes that jointly determine the target process through their causal relations. Streamflow, crucial for water resource management, is driven by meteorological variables through a series of interconnected processes. Key intermediate state variables, such as soil water and snowpack are regulated by precipitation, temperature, and basin characteristics. These state variables further influence the volume and timing of water entering rivers as streamflow.

HCMTL integrates the causal relationships among soil water, snowpack, and streamflow within a HMTnet. Soil water ($\text{SW}=\left\{\text{sw}_1,\text{sw}_2, \dots, \text{sw}_T \right\}$) and snowpack ($\text{SNO}=\left\{\text{sno}_1,\text{sno}_2, \dots, \text{sno}_T \right\}$) are treated as intermediate tasks, with their initial states ($\text{sw}_{\text{init}}$ and $\text{sno}_{\text{init}}$) and task embeddings ($h^{\text{sw}}$ and $h^{\text{sno}}$), along with basin characteristics and meteorological inputs $X=\left\{x_1,x_2, \dots, x_T \right\}$, supporting daily streamflow predictions $\text{SF}=\left\{\text{sf}_1,\text{sf}_2, \dots, \text{sf}_T \right\}$. 
Figure~\ref{fig:hcmtl} illustrates this HMTnet, comprising three separate LSTM modules for soil water (eq \ref{eq:sw}), snowpack (eq \ref{eq:sno}), and streamflow (eq \ref{eq:sf}), designed to predict each step i in a time series with a total of T steps.
\begin{align} \label{eq:sw}
\begin{split}
&p(\text{sw}_{i} | x_{i}, h^{\text{sw}}_{i-1}, \text{sw}_{\text{init}}) \\
&= \int p(\text{sw}_{i} | h^{\text{sw}}_{i}) p(h^{\text{sw}}_{i} | x_{i}, h^{\text{sw}}_{i-1}, \text{sw}_{\text{init}}) \, dh^{\text{sw}}_{i}
\end{split}
\vspace{-1\baselineskip}
\end{align}
\begin{align} \label{eq:sno}
\begin{split}
&p(\text{sno}_{i} | x_{i}, h^{\text{sno}}_{i-1}, \text{sno}_{\text{init}}) \\
&= \int p(\text{sno}_{i} | h^{\text{sno}}_{i}) p(h^{\text{sno}}_{i} | x_{i}, h^{\text{sno}}_{i-1}, \text{sno}_{\text{init}}) \, dh^{\text{sno}}_{i}
\end{split}
\vspace{-1\baselineskip}
\end{align}
\begin{align} \label{eq:sf}
\begin{split}
&p(\text{sf}_{i} | x_{i}, h^{\text{sf}}_{i-1}, h^{\text{sw}}_{i}, h^{\text{sno}}_{i}, \text{sf}_{\text{init}}) =\\
& \int p(\text{sf}_{i} | h^{\text{sf}}_{i}) p(h^{\text{sf}}_{i} | x_{i}, h^{\text{sf}}_{i-1}, h^{\text{sw}}_{i}, h^{\text{sno}}_{i}, \text{sf}_{\text{init}}) \, dh^{\text{sf}}_{i}
\end{split}
\vspace{-1\baselineskip}
\end{align}

HCMTL reduces task interference, enabling the model to capture variables that change at different temporal scales. The snowpack module can focus on its strong seasonal patterns, such as snow accumulation and melting, while the soil water module captures more immediate fluctuations driven by precipitation events. In the final stage, these intermediate modules are integrated into the target module, allowing the streamflow prediction to account for both short-term and long-term changes in soil water and snowpack.

\subsection{Embedding-Based Connections}
Due to their inherent complexity, most natural systems are not fully understood. Existing domain knowledge about their structure and processes is often incomplete and thus the physical representation of these systems are necessarily approximations of reality. For instance, while streamflow is influenced by soil water and snowpack, it also depends on other factors like evapotranspiration, canopy interception, agriculture irrigation, many of which are poorly understood and lack sufficient observations.

Existing approaches connect task modules using the output values of intermediate tasks ($\widehat{\text{sw}}$ and $\widehat{\text{sno}}$), which may provide incomplete approximations of reality. To address this, we propose using task embeddings ($h^{\text{sw}}$ and $h^{\text{sno}}$) to propagate information across different modules. Unlike output values, embeddings capture multi-dimensional internal states, offering more comprehensive representations of intermediate tasks. Additionally, embeddings are adaptable to unobserved processes, as they can encode other relevant processes beyond explicitly modeled task outputs. This helps to capture the complexities of hydrological processes more effectively.

\subsection{Conditional Mini-Batch (CMB) Learning}
Natural systems evolve over time, making it essential to uncover temporal patterns for a deeper understanding. However, existing temporal models (e.g., LSTM) often struggle with long time series (e.g., daily sequence over multiple years), requiring them to be divided into shorter segments. These models typically treat the segments independently, failing to capture their interactions and temporal dependencies, which results in suboptimal performance.

Figure \ref{fig:hcmtl} illustrates an example where a long time series with a total of T time steps, such as $\text{SF}=\left\{\text{sf}_1,\text{sf}_2, \dots, \text{sf}_T \right\}$, is divided into shorter segments, each with a length of t steps. HCMTL incorporates the CMB strategy to preserve segment interactions and capture cumulative changes, improving long time series modeling. 
During training, HCMTL uses the observed target values from one timestep before each segment (e.g., $\text{sf}_{\text{init}}$ in equation \ref{eq:sf}) as additional inputs, providing conditions for the model to learn cumulative changes for each time step $i$ within the segment. 
During inference, HCMTL maintains the temporal order of segments by connecting them using the predicted target values (e.g., $\widehat{\text{SF}}_{2t}$ in Figure \ref{fig:hcmtl}) from the current segment to initialize the next.
By integrating CMB, HCMTL more effectively retains segment interactions and long-term dependencies, uncovering temporal patterns in long time series from natural systems.

\section{Background} \label{sec:background}
\subsection{Hydrological cycle} \label{sec:hydro_cycle}
\begin{figure}
\centering
\includegraphics[width=\linewidth]{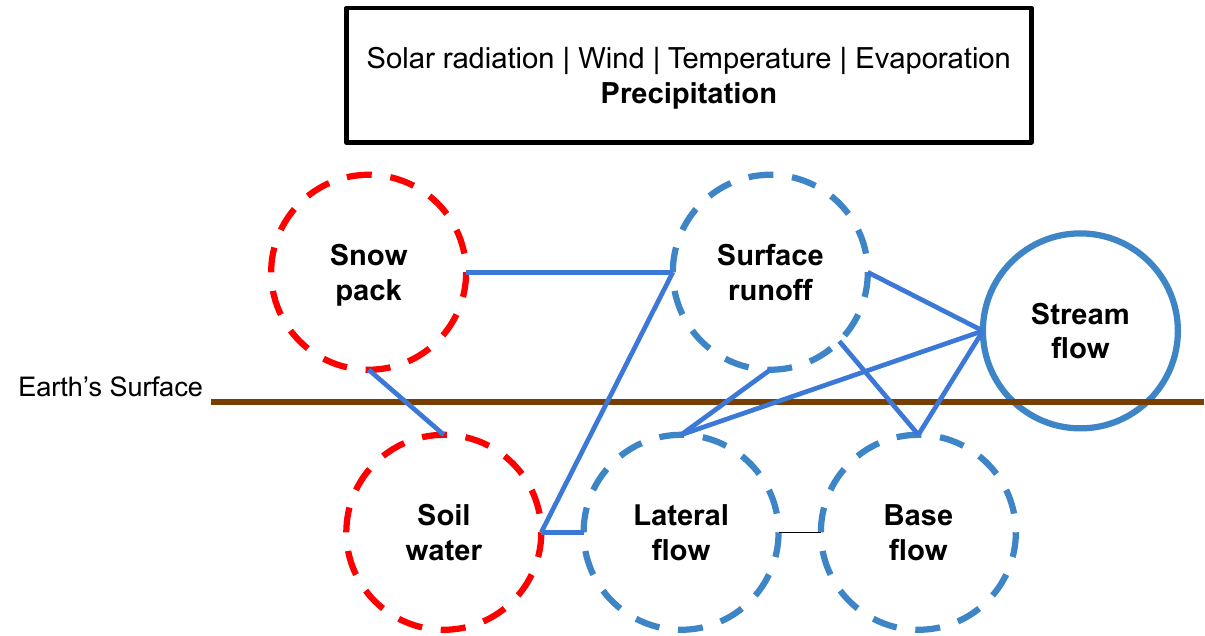}
\vspace{-1\baselineskip}
\caption{An abstraction of the hydrological cycle.}
\label{fig:sf_process}
\vspace{-0.5\baselineskip}
\end{figure}

Streamflow is connected to meteorological variables through a number of inter-connected processes as shown in Figure \ref{fig:sf_process}. Meteorological variables such as precipitation, temperature, evaporation, and solar radiation influence snowpack and soil water, which in turn affect streamflow. Precipitation contributes to snowpack when it falls as snow and increases soil water through infiltration when it falls as rain. Temperature and solar radiation melt the snowpack, generating flow, some of which may infiltrate into the subsurface and accumulate as soil water. Evapotranspiration, the process by which water is transferred from the land to the atmosphere through soil evaporation and plant transpiration, reduces soil water and can also cause snow sublimation, decreasing the snowpack. These intermediate target variables, snowpack and soil water, are crucial in determining the volume and timing of water entering streams. A massive snowpack, when melted in spring, produces a significant volume of runoff and boosts streamflow generation. Changes in soil moisture affect the dynamics of lateral flow and baseflow, both of which contribute to streamflow. Through these melting and subsurface processes, basin characteristics such as topography, soil type, and vegetation, along with climate attributes like seasonal patterns and variability, modulate these processes and ultimately determine the streamflow response.
\subsection{LSTM for modeling Dynamical Systems}
ML models that handle transformations from input sequences to targets are essential for extracting patterns from multivariate temporal data. Among these temporal models, RNNs are widespread using in diverse fields. A typical RNN can be viewed as an enhancement of traditional feed-forward networks, equipped with loops to process sequences of data. LSTM \cite{hochreiter1997long},  designed to prevent the vanishing gradient problem in standard RNNs, incorporate memory cells that help preserve relevant information throughout the sequence processing.
\begin{equation}
    \begin{split}
        \text{Forget gate}: \quad & f_t = \sigma (W_x^f x_t + W_h^f h_{t-1} + b^f )\\
        \text{Input gate}: \quad & i_t = \sigma (W_x^i x_t + W_h^i h_{t-1} + b^i )\\
        \text{Candidate}: \quad & \tilde{c}_t = \tanh (W_x^c x_t + W_h^c h_{t-1} + b^c )\\
        \text{Cell state}: \quad & c_t = f_t \odot c_{t-1} + i_t \odot \tilde{c}_t\\
        \text{Output gate}: \quad & o_t = \sigma (W_x^o x_t + W_h^o h_{t-1} + b^o )\\
        \text{Hidden state}: \quad & h_t = o_t \odot \tanh(c_t)
    \end{split}
    \vspace{-0.5\baselineskip}
\end{equation}
This setup operates in a many-to-many fashion. In regression, the model outputs $\hat{y}_t$ at each timestep $t$, computed as a linear function of the hidden states:
\begin{equation}
    \hat{y}_t = W_y h_t
    \vspace{-0.5\baselineskip}
\end{equation}
During training, long time series are divided into $K$ short segments of length $T$. The model minimizes mean squared error (MSE) loss across these segments. \ref{eq:mse_loss}.
\begin{equation}
    \mathcal{L} = \frac{1}{K} \sum_{k=1}^{K}\sum_{t=1}^{T} (y_t-\hat{y}_t)^2
    \label{eq:mse_loss}
    \vspace{-0.5\baselineskip}
\end{equation}

\subsection{Single-Task Learning (STL)}
STL trains a model to perform only one task. In hydrology, STL, particularly with LSTM networks, has been widely used to predict variables like water temperature \cite{jia2019physics, willard2022daily, zwart2023near} and water level \cite{bowes2019forecasting}. LSTM-based models have achieved state-of-the-art performance in streamflow modeling \cite{kratzert2019towards, kratzert2019ungauged} and flood forecasting \cite{yin2021rainfall, nearing2024global}. This success is largely due to advancements in large-sample hydrology\cite{addor2020large}, providing publicly available datasets from hundreds to thousands of basins. LSTM models trained on multi-basin data capture diverse rainfall–runoff responses, enhancing predictions by leveraging knowledge from large datasets.

Despite their effectiveness, STL models do not capture causal relationships between hydrological variables, potentially missing valuable information. They also struggle to generalize to out-of-sample cases due to task-specific specialization. Additionally, training separate models for related variables is inefficient. These challenges highlight the potential benefits of more advanced ML approaches, such as Multi-Task Learning.

\subsection{Multi-Task Learning (MTL)}
MTL is an advanced ML approach that trains a model to perform multiple related tasks by leveraging shared representations and interdependencies among the tasks. In hydrology, MTL improves model performance by capturing relationships between hydrological variables like streamflow, soil moisture, and evapotranspiration. By leveraging information across tasks, MTL enhances predictive accuracy and can be implemented as Simultaneous or Hierarchical MTL.

\subsubsection{Simultaneous Multi-Task Learning (SMTL)}
In SMTL, multiple related tasks are predicted concurrently using shared model layers. This enables the model to leverage interdependencies and shared representations among tasks.

In hydrology, \cite{sadler2022multi} applied SMTL to model daily streamflow and water temperature; \cite{ouyang2023exploring} and \cite{li2023improving} used it to improve evapotranspiration and streamflow predictions; and \cite{li2024enforcing} combined SMTL with physical constraints to forecast soil moisture, evapotranspiration, and runoff. These studies show that SMTL can match or surpass single-task models across multiple basins.

Despite its strengths, SMTL relies on task relatedness; if tasks are weakly correlated, shared representations can reduce performance.

\subsubsection{Hierachical Multi-Task Learning(HMTL)} HMTL uses separate network modules for different tasks, connecting them based on domain-specific causal relationships, making predictions more interpretable for scientists. HMTL minimizes interference by isolating unrelated tasks and enhances collaboration by grouping related ones. Its hierarchical design creates smaller networks for simpler tasks and larger networks for more complex tasks.

The concept of HMTL in hydrology was introduced by \cite{khandelwal2020physics} as a physics-guided deep learning architecture. While most hydrology studies still focus on SMTL, HMTL is emerging. In agriculture, HMTL has shown success : \cite{liu2022kgml} applied it to model N2O emissions, and \cite{liu2024knowledge} used it for carbon cycle modeling. Both studies found HMTL outperformed single-task models and state-of-the-art process-based models in their domains.

Existing HMTL models often connect task-specific modules using intermediate output values, overlooking underlying processes not reflected in domain causal knowledge. This reveals a research gap, which we address by using embeddings to connect network modules, as demonstrated in this paper.

\subsection{Mini-batch learning strategies}
To train machine learning models on long time series, it's common to divide the long time series into shorter segments, reducing computational complexity and mitigating vanishing/exploding gradient issues \cite{bengio1993problem,pascanu2013difficulty}. However, conventional learning algorithms often treat these segments as independent, lossing the interactions between segments and resulting in poorly trained models.

Conditional Mini-Batch Learning (CMB) \cite{xu2023mini} uses the initial response value of each segment as an additional input to preserve interactions between segments. This trains the model to capture cumulative changes within each segment, making it more robust against error accumulation compared to autoregressive methods like teacher forcing. CMB has shown improvements in STL multivariate time series modeling for variables such as soil moisture, snowpack, and streamflow \cite{xu2023mini,xu2024message}.
\section{Experiment Settings} \label{sec:settings}

\subsection{Data}
CARAVAN is a publicly available hydrology benchmark dataset for streamflow modeling~\cite{kratzert2022caravan}, consolidating and standardizing datasets from regions such as the United States, Great Britain, Central Europe, and more. Table ~\ref{tab:data_regions} presents the three regions and their basins with complete daily data from 1989 to 2009.

\begin{table}[htbp]
\caption{The number of basins in each region of interest.}
\vspace{-1\baselineskip}
\begin{center}
\resizebox{\linewidth}{!}{
\begin{tabular}{c|c c c}
\hline
&\textbf{CAMELS (US)}& \textbf{CAMELS-GB}&\textbf{LamaH-CE}\\
\hline
Region & United States&Great Britain&Central Europe\\
No. of basins & 319& 191& 250\\
\hline
\end{tabular}
}
\label{tab:data_regions}
\end{center}
\vspace{-0.5\baselineskip}
\end{table}
We split the data into a training set (1989–1996), a validation set (1997–1999), and a testing set (2000–2009). For each region, we calculate the regional mean and standard deviation from the training set and use them to normalize the training, validation, and testing sets. The input features include basin and climate attributes, derived from HydroATLAS ~\cite{linke2019global} and ERA5-Land ~\cite{munoz2021era5}, which remain static during the study period. Meteorological variables, also from ERA5-Land, are provided at a daily resolution.

We choose CARAVAN for because it is globally available, covering multiple regions, allowing us to test the hypothesis that a model performing well across regions is likely the best overall. Additionally, it provides soil water and snowpack time series, enabling the development of a HMTL networks based on their causal relationships with streamflow.
\subsection{Model Setup and Baselines}
\begin{figure}[htbp]
\centering
\includegraphics[width=\linewidth]{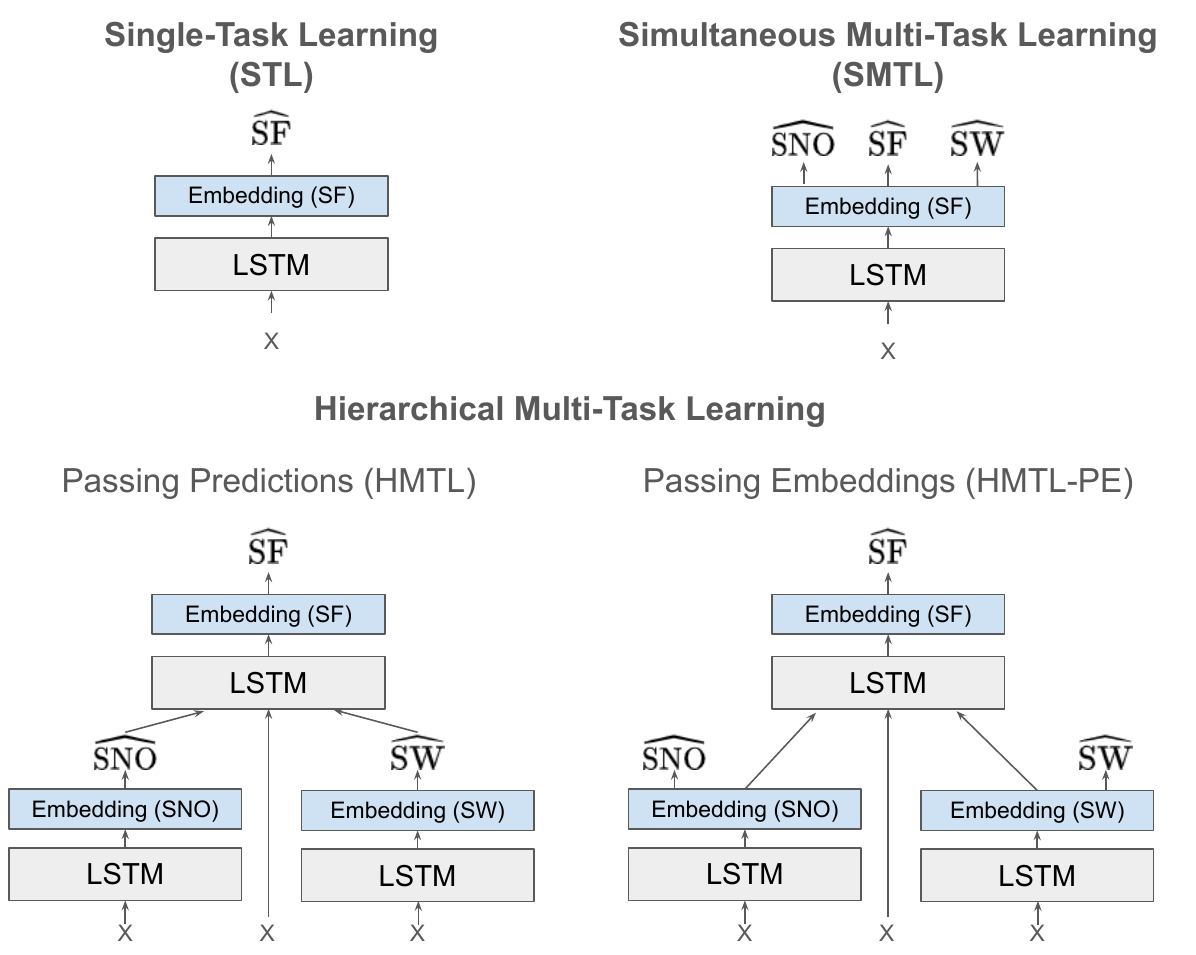}
\vspace{-1\baselineskip}
\caption{Model Setup and Baselines.}
\label{fig:arch}
\end{figure}

\begin{table*}[ht]
\caption{Summary of Benchmark Statistics for All Models Across Three Regions (Window: 365, Stride: 182)}
\vspace{-1\baselineskip}
\begin{center}
\resizebox{\linewidth}{!}{
\begin{tabular}{c|c c c|c c c|c c c}
\hline
\backslashbox{\textbf{Algo}}{\textbf{Data}}&\multicolumn{3}{c|}{\textbf{United States}}& \multicolumn{3}{c|}{\textbf{Great Britain}}& \multicolumn{3}{c}{\textbf{Austria}}\\
\hline
Streamflow 
 & $\downarrow$ RMSE & $\uparrow$\makecell{NSE\\mean} & $\uparrow$\makecell{NSE\\median}
 & $\downarrow$ RMSE & $\uparrow$\makecell{NSE\\mean} & $\uparrow$\makecell{NSE\\median}
 & $\downarrow$ RMSE & $\uparrow$\makecell{NSE\\mean} & $\uparrow$\makecell{NSE\\median}\\
\hline
STL \cite{kratzert2019towards}
&1.249&0.625&0.748
&1.287&0.665&0.695
&1.040&0.540&0.606\\
SMTL \cite{sadler2022multi}
&1.230&0.620&0.761
&1.250&0.683&0.713
&1.002&0.572&0.636\\
HMTL \cite{khandelwal2020physics}
&1.239&0.643&0.757
&1.271&0.676&0.699
&1.008&0.588&0.634\\
\hline
HMTL-CMB$_\text{ours}$
&1.228&0.645&0.765
&1.263&0.677&0.707
&0.996&0.592&0.637\\
HMTL-PE$_\text{ours}$
&1.229&0.650&\textbf{0.765}
&1.268&0.671&0.701
&0.994&\textbf{0.597}&0.638\\
HCMTL$_\text{ours}$
&\textbf{1.218}&\textbf{0.659}&0.764
&\textbf{1.248}&\textbf{0.689}&\textbf{0.716}
&\textbf{0.990}&0.594&\textbf{0.640}\\
\hline
\end{tabular}
}
\label{tab:overall_results}
\end{center}
\vspace{-1\baselineskip}
\end{table*}
All five baselines and the proposed HCMTL model use LSTMs as their core modules. The long time series is divided into sliding windows of 365 steps with a stride of 182, creating shorter segments for the models to process. Each LSTM module has a hidden state size of 256 units, with a dropout rate of 0.4 applied before a linear layer to generate the outputs.
\begin{itemize}
    \item STL: We use the LSTMs from \cite{kratzert2019towards, kratzert2019ungauged} to predict streamflow at each timestep. These studies showed that LSTMs outperform several CONUS-wide calibrated process-based models.
    \item SMTL: We use the Simultaneous Multi-Task LSTM model from \cite{sadler2022multi} to predict streamflow, soil water, and snowpack. \cite{sadler2022multi} showed that SMTL can improve accuracy by leveraging interdependent hydrologic variables.
    \item HMTL: We design the HMTL architecture based on \cite{khandelwal2020physics}, with separate LSTM modules for soil water and snowpack. Their predicted values are then used as additional inputs for the streamflow module to make predictions.
    \item HMTL-CMB: We integrated CMB \cite{xu2023mini} into HMTL to retain interactions between sample segments and capture cumulative changes within each segment.
    \item HMTL-PE: We enhanced HMTL by connecting task modules through passing embeddings (PE) instead of predicted values.
    \item HCMTL: The proposed HCMTL model enhances HMTL by connecting task modules through network embeddings and integrating CMB to retain segment interactions and learn cumulative changes.
\end{itemize}
\subsection{Training and Inference Procedures}
During training, the learning rate is set to 0.001, with Mean Square Error (MSE) as the loss function, equally weighted for each term. The batch size is 64, and the model is trained for 200 epochs, selecting the best model based on the lowest validation MSE. For each approach, six models with different initial weight configurations were trained.

During inference, we post-process the overlapping predicted segments to reconstruct the long time series. The latter part of each segment is used, as it tends to be more accurate by incorporating more time-step information. The final prediction is obtained by ensembling the predicted time series from the six trained models.

\subsection{Benchmark Metrics}
Root Mean Square Error (RMSE) is a commonly used metric in machine learning to assess prediction accuracy. While intuitive, RMSE can be skewed by large errors in high-streamflow basins. In this study, we calculate the RMSE for each basin and then average these RMSEs to obtain the final RMSE.
\begin{equation}
    \text{RMSE} = \sqrt{\sum_{t=1}^{T} (y_t-\hat{y}_t)^2}
    \label{eq:rmse}
    \vspace{-0.5\baselineskip}
\end{equation}

The Nash-Sutcliffe Efficiency (NSE) is a commonly used metric in hydrology, comparing model performance to a mean-based baseline. NSE values range from 1 (perfect match) to negative values (indicating performance worse than the mean model). In this study, we calculate the mean baseline and NSE for each basin, then average these NSEs to obtain a final NSE. This serves as a normalized measure for comparing basins with significantly different streamflow volumes. However, the final NSE can be skewed by highly negative values in low-streamflow basins. Thus, NSE complements RMSE, offering a more complete understanding of model performance.
\begin{equation}
    \text{NSE} = 1 - \frac{\sum_{t=1}^{T} \left( y_t - \hat{y}_t \right)^2}{\sum_{t=1}^{T} \left( y_t - \bar{y} \right)^2}
    \label{eq:nse}
    \vspace{-0.5\baselineskip}
\end{equation}

\begin{figure}
    \centering
    \includegraphics[width=\linewidth]{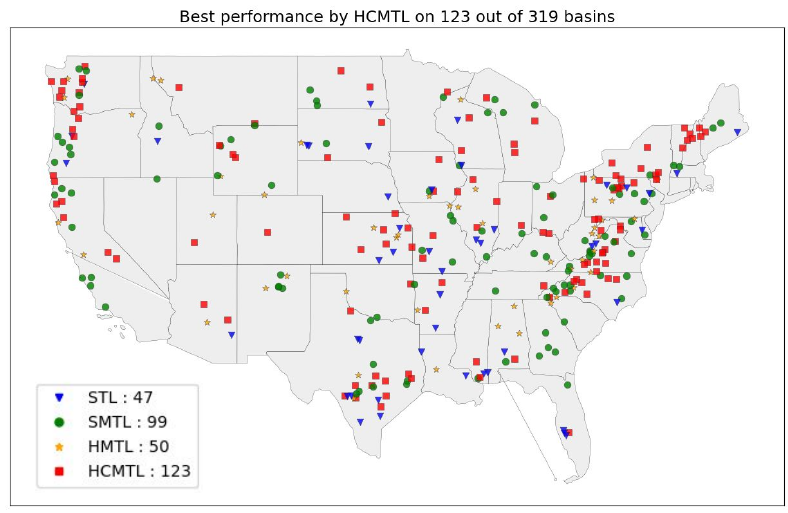}
    \vspace{-1\baselineskip}
    \caption{HCMTL achieves the best performance in 123 out of 319 basins in the United States.}
    \label{fig:map_usa}
\end{figure}
\section{Results} \label{sec:results}
\subsection{The overall performance}
Table \ref{tab:overall_results} shows that the proposed HCMTL model delivers the best overall performance, with lower RMSEs and higher NSEs. HCMTL consistently achieves the lowest RMSE across all four regions, highlighting its superior accuracy. We conduct a detailed study of the United States in the following sections. Figure \ref{fig:map_usa} shows HCMTL performing best in 123 out of 319 U.S. basins.

\begin{figure}
    \centering
    \includegraphics[width=\linewidth]{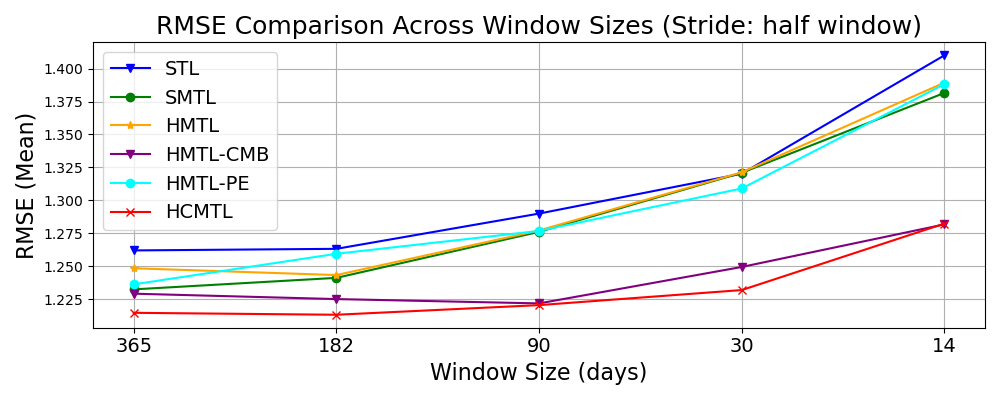}
    \vspace{-1\baselineskip}
    \caption{All models show higher RMSEs as segment length decreases, as shorter segments capture less temporal information and result in a greater loss of interactions between segments.}
    \label{fig:winsize_usa}
\end{figure}
\subsection{Effect of Segment Length} \label{sec:segment_length}
To train machine learning models on long time series, it is common to divide them into shorter segments to reduce computational complexity and mitigate vanishing or exploding gradient issues \cite{pascanu2013difficulty, wen2022transformers}. To investigate the impact of segment length on model performance, we implemented a sliding window approach with window sizes of 365, 182, 90, 30, and 14 days, using a stride of half the window size. For each of the 319 US basins, a 14-day window produces 364 segments per basin, while a 365-day window generates 14 segments per basin.

Figure \ref{fig:winsize_usa} shows that HCMTL consistently outperforms the baselines, especially as the window size decreases. This is due to HCMTL's use of CMB, which preserves interactions and dependencies between segments during training and inference.

\subsection{Effect of Stride Length} \label{appendix:stride}
To reduce the loss of temporal information in long time series, we can implement a sliding window approach with a fixed window size and smaller strides to introduce more overlap between segments. In this experiment, we fixed the window size at 365 days and varied the strides from 365 to 90 days. The 365-day stride starts each segment on the first day of the water year, while the 90-day stride starts each segment at the beginning of each season, thus retaining more granular temporal information.

\begin{figure}
    \centering
    \includegraphics[width=\linewidth]{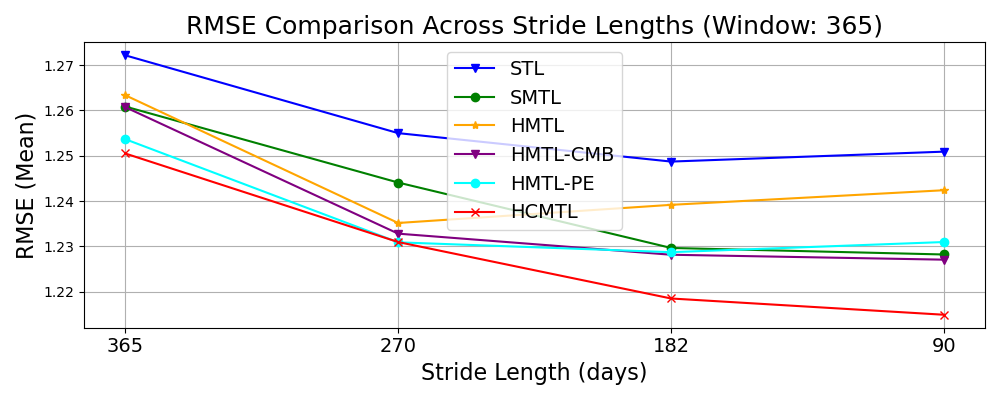}
    \vspace{-1\baselineskip}
    \caption{HCMTL outperforms baselines, with the performance gap widening as stride length decreases.}
    \label{fig:stride_length}
\end{figure}

Figure \ref{fig:stride_length} shows that models achieve lower RMSEs as the stride decreases, which increases the overlap between segments and retains more temporal information. HCMTL consistently outperforms the baselines, with the performance gap widening as the stride length decreases. This is because smaller strides create more segments, increasing the number of segment interactions. While other models struggle to maintain these interactions, HCMTL effectively preserves them. Additionally, HCMTL connects network modules using embeddings, making the model more expressive for larger datasets. Both the CMB and embedding design choices contribute to HCMTL's superior performance.
\subsection{Effect of Training Data Size} \label{sec:datasize}
\begin{figure}
    \centering
    \includegraphics[width=\linewidth]{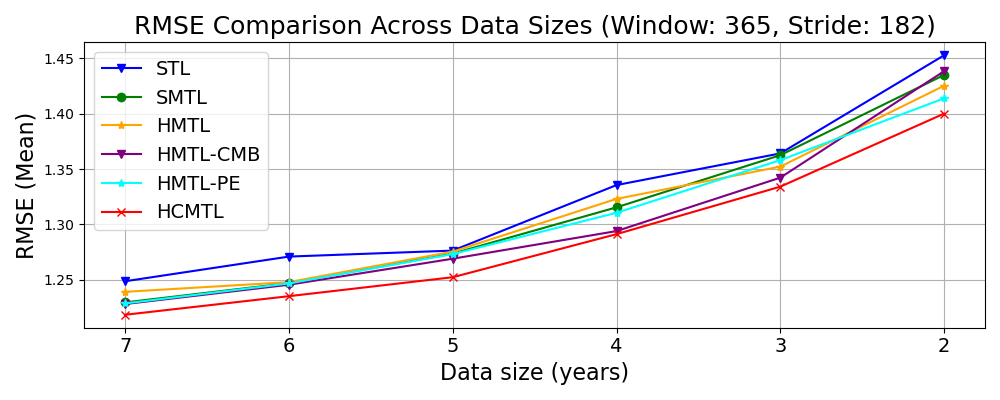}
    \vspace{-1\baselineskip}
    \caption{All models show higher RMSEs as training data size decrease.}
    \label{fig:datasize_usa}
    \vspace{-0.3\baselineskip}
\end{figure}

In hydrology, developing models that perform reliably in data-scarce situations is essential due to the high cost of collecting streamflow data. This experiment evaluates model robustness using a 365-day window size, a 182-day stride, and training data sizes of 7, 6, 5, 4, 3, and 2 years. Figure \ref{fig:datasize_usa} shows that all models experience higher RMSEs as training data size decreases, with HCMTL consistently achieving the lowest RMSEs.

\subsection{Effect of Noise on Intermediate Targets}
Testing models under varying noise levels is crucial for assessing their robustness and reliability, as real-world data is seldom perfect. In this experiment, Gaussian noise scaled to each basin's standard deviation (levels of 0, 0.01, 0.1, 0.5, 1, and 2) was added at each time step to corrupt the soil water and snowpack time series.

\begin{figure}
    \centering
    \includegraphics[width=\linewidth]{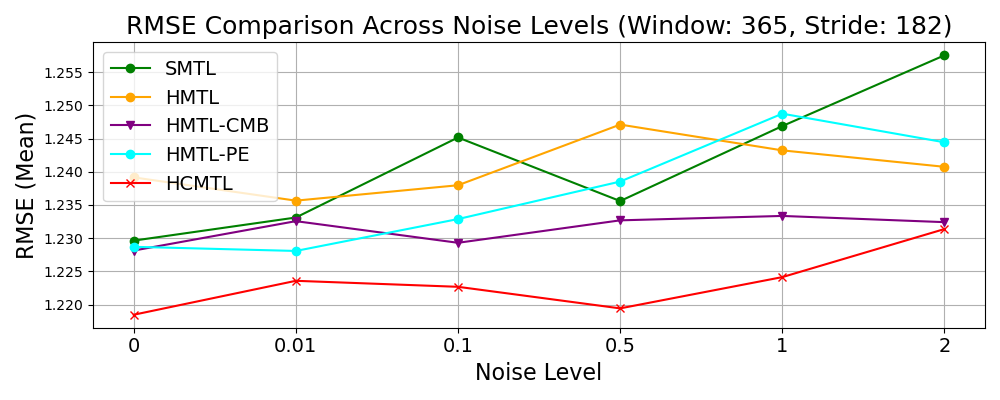}
    \vspace{-1\baselineskip}
    \caption{Models show higher RMSEs as noise increases in the soil water and snowpack time series, reducing their effectiveness in streamflow modeling. HCMTL and HCMTL-CMB demonstrate greater robustness to noise.}
    \label{fig:noise_usa}
\end{figure}

Figure \ref{fig:noise_usa} shows that HCMTL achieves the lowest RMSEs until noise levels exceed 0.5, after which its RMSEs begin to rise. In contrast, HMTL-CMB’s RMSEs remain stable as noise levels increase. A similar trend is observed in HMTL-PE, which consistently outperforms HMTL until noise levels exceed 1. These suggests that passing predicted values is more resilient to high noise than passing embeddings between task modules. Predicted values introduce only two noisy channels, which the network can easily ignore, while embeddings add 256 noisy channels, making it harder to filter out noise. However, entirely corrupted channels should be removed during preprocessing. In practice, input channels often have low to moderate noise levels, where HCMTL performs best, demonstrating its practical value.

\section{Discussion} \label{sec:discussion}
\subsection{Impact of Joint Modeling} The results show that SMTL consistently outperforms STL, even though both use the same LSTM model. This underscores the advantages of joint modeling with soil water and snowpack for improving streamflow predictions.

\subsection{Hierarchical Architectures}
Table \ref{tab:overall_results} shows that HMTL generally underperforms compared to SMTL models, while HMTL-PE outperforms both. This suggests that using embeddings is more effective than predicted values for connecting task modules in hierarchical networks. 
Figures \ref{fig:winsize_usa}, \ref{fig:stride_length}, and \ref{fig:datasize_usa} further demonstrate that HMTL-PE outperforms HMTL across various sliding window sizes, strides, and training data sizes, indicating greater robustness.
Figure \ref{fig:noise_usa} shows that HMTL-PE outperforms HMTL when small to moderate noise is injected into the soil water and snowpack time series. HMTL only surpasses HMTL-PE under conditions of high noise, where the time series are entirely corrupted. In practice, such corrupted channels are typically removed during preprocessing, and models generally deal with small to moderate noise. Therefore, we consider HMTL-PE the better approach.

\subsection{Integrating Conditional Mini-Batch Learning (CMB)}
Table \ref{tab:overall_results} shows that HMTL-CMB outperforms HMTL in the United States, Great Britain, and Austria. Figures \ref{fig:winsize_usa} and \ref{fig:stride_length} demonstrate that the performance gap widens as sliding window sizes or strides decrease in the U.S. data. Figure \ref{fig:datasize_usa} further shows that HMTL-CMB continues to outperform HMTL until the training time series is reduced to two years, producing only three segments per basin (365-day segment length, 182-day stride), where CMB becomes less effective due to limited segment interactions.  Additionally, Figure \ref{fig:noise_usa} shows that HMTL-CMB maintains its performance under high noise levels, outperforming HMTL and demonstrating its robustness. These results highlight that CMB enhances performance by preserving segment interactions and modeling cumulative changes, proving effective for long time series modeling.

\subsection{Hierarchical Conditional Multi-Task Learning (HCMTL)} The results show that HCMTL often outperforms HMTL-CMB, highlighting the advantage of connecting task modules through embeddings. Similarly, HCMTL frequently performs better than HMTL-PE, demonstrating the value of integrating CMB into the architecture. These findings indicate that the two design choices complement each other, jointly contributing to the superior performance of HCMTL in streamflow modeling across regions. 
\section{Conclusions}
We propose Hierarchical Conditional Multi-Task Learning (HCMTL), a hierarchical approach that models soil water and snowpack as intermediate tasks based on their causal links to streamflow. HCMTL uses task embeddings to connect network modules, enhancing flexibility and capturing unobserved processes. It also incorporates a Conditional Mini-Batch strategy to improve long time series modeling. HCMTL outperforms five baselines across hundreds of basins in the United States, Great Britain, and Austria. 
We further conducted comprehensive experiments in the United States—reducing window size, stride, and training data, while increasing noise in intermediate tasks—and confirmed the robustness of HCMTL in streamflow modeling.

Future work could integrate more intermediate processes, such as surface runoff and lateral flow, and apply mass conservation-based loss functions for physics-consistent predictions. Advanced algorithms \cite{xu2024message} could also be integrated to enhance long time series modeling. Lastly, HCMTL’s approach is adaptable to other complex systems, such as healthcare and finance, where multiple processes influence outcomes over long time scales.

\bibliographystyle{unsrt}
\bibliography{references}

\begin{thebibliography}{10}

\bibitem{bieger2017introduction}
Katrin Bieger, Jeffrey~G Arnold, Hendrik Rathjens, Michael~J White, David~D Bosch, Peter~M Allen, Martin Volk, and Raghavan Srinivasan.
\newblock Introduction to swat+, a completely restructured version of the soil and water assessment tool.
\newblock {\em JAWRA Journal of the American Water Resources Association}, 53(1):115--130, 2017.

\bibitem{saad2019estimates}
David~A Saad, Gregory~E Schwarz, Denise~M Argue, David~W Anning, Scott~A Ator, Anne~B Hoos, Stephen~D Preston, Dale~M Robertson, and Daniel Wise.
\newblock Estimates of long-term mean daily streamflow and annual nutrient and suspended-sediment loads considered for use in regional sparrow models of the conterminous united states, 2012 base year.
\newblock Technical report, US Geological Survey, 2019.

\bibitem{addor2020large}
Nans Addor, Hong~X Do, Camila Alvarez-Garreton, Gemma Coxon, Keirnan Fowler, and Pablo~A Mendoza.
\newblock Large-sample hydrology: recent progress, guidelines for new datasets and grand challenges.
\newblock {\em Hydrological Sciences Journal}, 65(5):712--725, 2020.

\bibitem{kratzert2019towards}
Frederik Kratzert, Daniel Klotz, Guy Shalev, G{\"u}nter Klambauer, Sepp Hochreiter, and Grey Nearing.
\newblock Towards learning universal, regional, and local hydrological behaviors via machine learning applied to large-sample datasets.
\newblock {\em Hydrology and Earth System Sciences}, 23(12):5089--5110, 2019.

\bibitem{kratzert2019ungauged}
Frederik Kratzert, Daniel Klotz, Mathew Herrnegger, Alden~K Sampson, Sepp Hochreiter, and Grey~S Nearing.
\newblock Toward improved predictions in ungauged basins: Exploiting the power of machine learning.
\newblock {\em Water Resources Research}, 55(12):11344--11354, 2019.

\bibitem{gauch2021rainfall}
Martin Gauch, Frederik Kratzert, Daniel Klotz, Grey Nearing, Jimmy Lin, and Sepp Hochreiter.
\newblock Rainfall--runoff prediction at multiple timescales with a single long short-term memory network.
\newblock {\em Hydrology and Earth System Sciences}, 25(4):2045--2062, 2021.

\bibitem{nearing2024global}
Grey Nearing, Deborah Cohen, Vusumuzi Dube, Martin Gauch, Oren Gilon, Shaun Harrigan, Avinatan Hassidim, Daniel Klotz, Frederik Kratzert, Asher Metzger, et~al.
\newblock Global prediction of extreme floods in ungauged watersheds.
\newblock {\em Nature}, 627(8004):559--563, 2024.

\bibitem{bengio1993problem}
Yoshua Bengio, Paolo Frasconi, and Patrice Simard.
\newblock The problem of learning long-term dependencies in recurrent networks.
\newblock In {\em IEEE international conference on neural networks}, pages 1183--1188. IEEE, 1993.

\bibitem{pascanu2013difficulty}
Razvan Pascanu, Tomas Mikolov, and Yoshua Bengio.
\newblock On the difficulty of training recurrent neural networks.
\newblock In {\em International conference on machine learning}, pages 1310--1318. Pmlr, 2013.

\bibitem{xu2023mini}
Shaoming Xu, Ankush Khandelwal, Xiang Li, Xiaowei Jia, Licheng Liu, Jared Willard, Rahul Ghosh, Kelly Cutler, Michael Steinbach, Christopher Duffy, et~al.
\newblock Mini-batch learning strategies for modeling long term temporal dependencies: A study in environmental applications.
\newblock In {\em Proceedings of the 2023 SIAM International Conference on Data Mining (SDM)}, pages 649--657. SIAM, 2023.

\bibitem{kratzert2022caravan}
Frederik Kratzert, Grey Nearing, Nans Addor, Tyler Erickson, Martin Gauch, Oren Gilon, Lukas Gudmundsson, Avinatan Hassidim, Daniel Klotz, Sella Nevo, et~al.
\newblock Caravan-a global community dataset for large-sample hydrology.
\newblock {\em Scientific Data}, 10(1):61, 2023.

\bibitem{hochreiter1997long}
Sepp Hochreiter and J{\"u}rgen Schmidhuber.
\newblock Long short-term memory.
\newblock {\em Neural computation}, 9(8):1735--1780, 1997.

\bibitem{jia2019physics}
Xiaowei Jia, Jared Willard, Anuj Karpatne, Jordan Read, Jacob Zwart, Michael Steinbach, and Vipin Kumar.
\newblock Physics guided rnns for modeling dynamical systems: A case study in simulating lake temperature profiles.
\newblock In {\em Proceedings of the 2019 SIAM international conference on data mining}, pages 558--566. SIAM, 2019.

\bibitem{willard2022daily}
Jared~D Willard, Jordan~S Read, Simon Topp, Gretchen~JA Hansen, and Vipin Kumar.
\newblock Daily surface temperatures for 185,549 lakes in the conterminous united states estimated using deep learning (1980--2020).
\newblock {\em Limnology and Oceanography Letters}, 7(4):287--301, 2022.

\bibitem{zwart2023near}
Jacob~A Zwart, Samantha~K Oliver, William~David Watkins, Jeffrey~M Sadler, Alison~P Appling, Hayley~R Corson-Dosch, Xiaowei Jia, Vipin Kumar, and Jordan~S Read.
\newblock Near-term forecasts of stream temperature using deep learning and data assimilation in support of management decisions.
\newblock {\em JAWRA Journal of the American Water Resources Association}, 59(2):317--337, 2023.

\bibitem{bowes2019forecasting}
Benjamin~D Bowes, Jeffrey~M Sadler, Mohamed~M Morsy, Madhur Behl, and Jonathan~L Goodall.
\newblock Forecasting groundwater table in a flood prone coastal city with long short-term memory and recurrent neural networks.
\newblock {\em Water}, 11(5):1098, 2019.

\bibitem{yin2021rainfall}
Hanlin Yin, Xiuwei Zhang, Fandu Wang, Yanning Zhang, Runliang Xia, and Jin Jin.
\newblock Rainfall-runoff modeling using lstm-based multi-state-vector sequence-to-sequence model.
\newblock {\em Journal of Hydrology}, 598:126378, 2021.

\bibitem{sadler2022multi}
Jeffrey~Michael Sadler, Alison~Paige Appling, Jordan~S Read, Samantha~Kay Oliver, Xiaowei Jia, Jacob~Aaron Zwart, and Vipin Kumar.
\newblock Multi-task deep learning of daily streamflow and water temperature.
\newblock {\em Water Resources Research}, 58(4):e2021WR030138, 2022.

\bibitem{ouyang2023exploring}
Wenyu Ouyang, Xuezhi Gu, Lei Ye, Xiaoning Liu, and Chi Zhang.
\newblock Exploring variable synergy in multi-task deep learning for hydrological modeling.
\newblock {\em Authorea Preprints}, 2023.

\bibitem{li2023improving}
Bu~Li, Ruidong Li, Ting Sun, Aofan Gong, Fuqiang Tian, Mohd Yawar~Ali Khan, and Guangheng Ni.
\newblock Improving lstm hydrological modeling with spatiotemporal deep learning and multi-task learning: A case study of three mountainous areas on the tibetan plateau.
\newblock {\em Journal of Hydrology}, 620:129401, 2023.

\bibitem{li2024enforcing}
Lu~Li, Yongjiu Dai, Zhongwang Wei, Wei Shangguan, Yonggen Zhang, Nan Wei, and Qingliang Li.
\newblock Enforcing water balance in multitask deep learning models for hydrological forecasting.
\newblock {\em Journal of Hydrometeorology}, 25(1):89--103, 2024.

\bibitem{khandelwal2020physics}
Ankush Khandelwal, Shaoming Xu, Xiang Li, Xiaowei Jia, Michael Stienbach, Christopher Duffy, John Nieber, and Vipin Kumar.
\newblock Physics guided machine learning methods for hydrology.
\newblock {\em arXiv preprint arXiv:2012.02854}, 2020.

\bibitem{liu2022kgml}
Licheng Liu, Shaoming Xu, Jinyun Tang, Kaiyu Guan, Timothy~J Griffis, Matthew~D Erickson, Alexander~L Frie, Xiaowei Jia, Taegon Kim, Lee~T Miller, et~al.
\newblock Kgml-ag: a modeling framework of knowledge-guided machine learning to simulate agroecosystems: a case study of estimating n 2 o emission using data from mesocosm experiments.
\newblock {\em Geoscientific model development}, 15(7):2839--2858, 2022.

\bibitem{liu2024knowledge}
Licheng Liu, Wang Zhou, Kaiyu Guan, Bin Peng, Shaoming Xu, Jinyun Tang, Qing Zhu, Jessica Till, Xiaowei Jia, Chongya Jiang, et~al.
\newblock Knowledge-guided machine learning can improve carbon cycle quantification in agroecosystems.
\newblock {\em Nature communications}, 15(1):357, 2024.

\bibitem{xu2024message}
Shaoming Xu, Ankush Khandelwal, Arvind Renganathan, and Vipin Kumar.
\newblock Message propagation through time: An algorithm for sequence dependency retention in time series modeling.
\newblock In {\em Proceedings of the 2024 SIAM International Conference on Data Mining (SDM)}, pages 307--315. SIAM, 2024.

\bibitem{linke2019global}
Simon Linke, Bernhard Lehner, Camille Ouellet~Dallaire, Joseph Ariwi, G{\"u}nther Grill, Mira Anand, Penny Beames, Vicente Burchard-Levine, Sally Maxwell, Hana Moidu, et~al.
\newblock Global hydro-environmental sub-basin and river reach characteristics at high spatial resolution.
\newblock {\em Scientific data}, 6(1):283, 2019.

\bibitem{munoz2021era5}
Joaqu{\'\i}n Mu{\~n}oz-Sabater, Emanuel Dutra, Anna Agust{\'\i}-Panareda, Cl{\'e}ment Albergel, Gabriele Arduini, Gianpaolo Balsamo, Souhail Boussetta, Margarita Choulga, Shaun Harrigan, Hans Hersbach, et~al.
\newblock Era5-land: A state-of-the-art global reanalysis dataset for land applications.
\newblock {\em Earth system science data}, 13(9):4349--4383, 2021.

\bibitem{wen2022transformers}
Qingsong Wen, Tian Zhou, Chaoli Zhang, Weiqi Chen, Ziqing Ma, Junchi Yan, and Liang Sun.
\newblock Transformers in time series: A survey.
\newblock {\em arXiv preprint arXiv:2202.07125}, 2022.

\bibitem{newman2014large}
AJ~Newman, K~Sampson, MP~Clark, A~Bock, RJ~Viger, and D~Blodgett.
\newblock A large-sample watershed-scale hydrometeorological dataset for the contiguous usa.
\newblock {\em Boulder, CO: UCAR/NCAR}, 2014.

\bibitem{addor2017large}
Nans Addor, Andrew~J. Newman, Naoki Mizukami, and Martyn~P. Clark.
\newblock Catchment attributes for large-sample studies.
\newblock {\em Boulder, CO: UCAR/NCAR}, 2017.

\bibitem{wood2002long}
Andrew~W Wood, Edwin~P Maurer, Arun Kumar, and Dennis~P Lettenmaier.
\newblock Long-range experimental hydrologic forecasting for the eastern united states.
\newblock {\em Journal of Geophysical Research: Atmospheres}, 107(D20):ACL--6, 2002.

\end{thebibliography}
\clearpage
\newpage
\appendix
\section{Appendix}
\begin{figure*}[htbp]
\centering
\includegraphics[width=\linewidth]{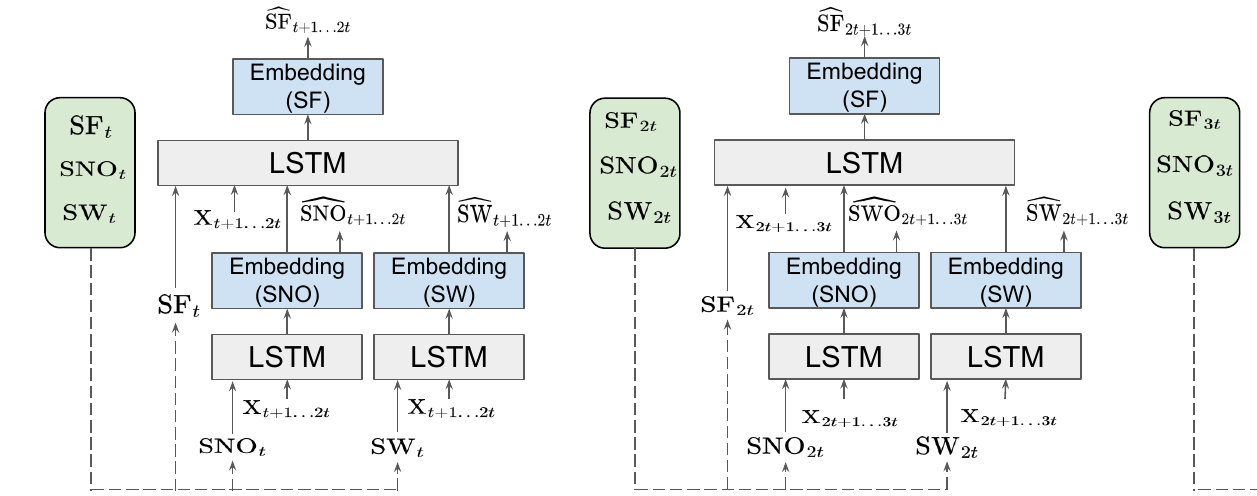}
\vspace{-1\baselineskip}
\caption{Hierarchical Conditional Multi-Task Learning for streamflow (SF) modeling during training.}
\label{fig:hcmtl_training}
\vspace{-1\baselineskip}
\end{figure*}

\subsection{HCMTL Training Algorithm}
Figure \ref{fig:hcmtl_training} shows that the HCMTL training algorithm uses the observed target values from one timestep before each segment as additional inputs, providing conditions for the model to learn cumulative changes for each time step $i$ within the segment. Since HCMTL can directly retrieve these initial target values from the training data, the HCMTL training algorithm allows random shuffling of segments to create mini-batches for model training. Therefore, the HCMTL training algorithm can be considered as a mini-batch gradient descent method, enhanced by initial target values to learn cumulative changes within segments and preserve interactions between segments.

\subsection{CARAVAN Dataset}
CARAVAN is a publicly available hydrology benchmark dataset for streamflow modeling~\cite{kratzert2022caravan}, consolidating and standardizing data from regions including the United States, Great Britain, Central Europe, and others.

The input features include basin and climate attributes, as well as meteorological variables. The basin and climate attributes, derived from HydroATLAS ~\cite{linke2019global} and ERA5-Land ~\cite{munoz2021era5}, remain static throughout the study period, while meteorological variables from ERA5-Land are provided at a daily resolution. We select a subset of CARAVAN input features (Tables ~\ref{tab:basin_attributes} and ~\ref{tab:met_variables}) that align with those in CAMELS, enabling researchers to apply the proposed architecture to the CAMELS dataset. Although CARAVAN’s meteorological variables are generally less accurate (i.e., more uncertain) than high-resolution, localized datasets like CAMELS~\cite{kratzert2022caravan}, we choose CARAVAN for two key reasons. First, it is globally available and spans multiple regions, allowing us to verify the hypothesis that an architecture performing well across diverse regions is likely the best overall. Second, it includes time series data for modeled soil water and snowpack from ERA5-Land (Table ~\ref{tab:state_variables}), enabling the incorporation of hydrological states into the architecture for streamflow modeling.
\begin{table}[htbp]
\caption{Summary of basin and climate attributes.}
\resizebox{\columnwidth}{!}{%
\begin{tabular}{|l|l|l|}
\hline
\textbf{Variable Name} & \textbf{Description} & \textbf{Unit}                                                                                                                                      \\ \hline
p\_mean                & Mean daily precipitation &  $mm/day$                                                                                                                                                \\ \hline
pet\_mean              & Mean daily potential evaporation &  $mm/day$                                                                                                                                          \\ \hline
aridity                & \begin{tabular}[c]{@{}l@{}}Aridity index, ratio of \\mean PET and mean precipitation\end{tabular} & --                                                                                                                   \\ \hline
frac\_snow             & Fraction of precipitation falling as snow     & --                                                                                                                             \\ \hline
moisture\_index        & Mean annual moisture index in range {[}-1, 1{]}  &--                                                                                                                        \\ \hline
seasonality            & Moisture index seasonality in range {[}0, 2{]} &--                                                                                                                            \\ \hline
kar\_pc\_sse           & Karst area extent    & \% cover                                                                                                                                                      \\ \hline
cly\_pc\_sav           & Clay fraction in soil    & \%                                                                                                                                                  \\ \hline
slt\_pc\_sav           & Silt fraction in soil   & \%                                                                                                                                                   \\ \hline
snd\_pc\_sav           & Sand fraction in soil  & \%                                                                                                                                                    \\ \hline
soc\_th\_sav           & Organic carbon content in soil  & tonnes/hectare                                                                                                                                           \\ \hline
swc\_pc\_syr           & Annual mean soil water content       & \%                                                                                                                                      \\ \hline
ele\_mt\_sav           & Elevation     & $m$ above sea level                                                                                                                                                             \\ \hline
slp\_dg\_sav           & Terrain slope  & °(x10)                                                                                                                                                            \\ \hline
basin\_area            & Basin Area   &      $km^2$                                                                                                                                                        \\ \hline
for\_pc\_sse           & Forest cover extent    & \% cover                                                                                                                                                    \\ \hline
\end{tabular}%
}
\label{tab:basin_attributes}
\vspace{-0.8\baselineskip}
\end{table}

\begin{table}[htbp]
    \caption{ Summary of meteorological variables.}
    \resizebox{\linewidth}{!}{
        \begin{tabular}{|l|l|l|}
        \hline
        \textbf{Meteorological forcing data}                                               &\textbf{Description} & \textbf{Unit} \\ \hline
        total\_precipitation\_sum                                 &Daily Precipitation sum& $mm/day$        \\ \hline %
        potential\_evaporation\_sum                       &Daily Potential evaporation sum& $mm/day$        \\ \hline %
        temperature\_2m\_mean                                       &Mean Air temperature& $^\circ C$            \\ \hline %
        dewpoint\_temperature\_2m\_mean                       &Mean Dew point temperature& $^\circ C$            \\ \hline %
        surface\_net\_solar\_radiation\_mean                    &Mean Shortwave radiation& $W/m^{2}$          \\ \hline %
        surface\_net\_thermal\_radiation\_mean &Mean Net thermal radiation at the surface& $W/m^{2}$          \\ \hline %
        surface\_pressure\_mean                                    &Mean Surface pressure& $kPa$           \\ \hline %
        u\_component\_of\_wind\_10m\_mean                   &Mean Eastward wind component& $m/s$          \\ \hline %
        v\_component\_of\_wind\_10m\_mean                  &Mean Northward wind component& $m/s$         \\ \hline %
        \end{tabular}
    }
    \label{tab:met_variables}
\vspace{-0.5\baselineskip}
\end{table}

\begin{table}[htbp]
    \caption{Summary of target variables.}
    \resizebox{\linewidth}{!}{
        \begin{tabular}{|l|l|l|}
        \hline
        \textbf{State and target variables.}                                               &\textbf{Description}& \textbf{Unit}      \\ \hline
        snow\_depth\_water\_equivalent                                 &Snow water equivalent & $mm$           \\ \hline
        volumetric\_soil\_water\_layer\_1                      &Soil water volume 0–7 cm& $m^3/m^3$           \\ \hline
        Streamflow                                       &Streamflow& $mm/day$         \\ \hline
        \end{tabular}
    }
    \label{tab:state_variables}
\vspace{-0.5\baselineskip}
\end{table}

\subsection{CAMELS Dataset}
The Catchment Attributes and Meteorological (CAMELS) dataset \cite{newman2014large, addor2017large}, curated by the US National Center for Atmospheric Research (NCAR), includes the basins analyzed in this study. CAMELS uses daily basin-averaged Maurer forcings \cite{wood2002long} for time-dependent meteorological inputs, which are of higher quality than those provided by CARAVAN. To leverage this higher-resolution data, we supplement CARAVAN's meteorological inputs (Table \ref{tab:met_variables}) with five additional CAMELS variables (Table \ref{tab:met_variables_camels}) when training models for the United States.

\begin{table}[htbp]
    \caption{Summary of CAMELS's meteorological variables.}
    \resizebox{\linewidth}{!}{
        \begin{tabular}{|l|l|l|}
        \hline
        \textbf{Meteorological forcing data}                                               &\textbf{Description} & \textbf{Unit} \\ \hline
        camels\_PRCP                                 &Daily Precipitation sum& $mm/day$        \\ \hline %
        camels\_SRAD                    &Mean Shortwave radiation& $W/m^2$          \\ \hline %
        camels\_Tmax                                      &Maximum Air temperature& $^\circ C$            \\ \hline %
        camels\_Tmin                                      &Minimum Air temperature& $^\circ C$            \\ \hline %
        camels\_Vp                                    &Vapor pressure& $Pa$           \\ \hline %
        \end{tabular}
    }
    \label{tab:met_variables_camels}
\vspace{-0.5\baselineskip}
\end{table}

\subsection{Computational Complexity}
\begin{table}[htbp]
    \caption{Computational Complexity}
    \label{tab:computational_complexity}
    \resizebox{\linewidth}{!}{
    \begin{tabular}{c|c|c|c|c|c}
        \multicolumn{6}{c}{USA, 7 Years, Window: 365, Stride: 182, hidden states: 256}\\
        \hline
        Metric & Unit & STL & SMTL & HMTL & HCMTL\\
        \hline
        Parameters & count &306433&306947&921347&941586\\
        Training Time  & seconds/epoch &11.0082&11.4506&32.1375&31.5727\\
        Inference Time & seconds &7.61&8.99&9.22&8.99\\
        \hline
    \end{tabular}
    }
\vspace{-0.5\baselineskip}
\end{table}
Table \ref{tab:computational_complexity} shows that HMTL and HCMTL have a similar number of neural network parameters, demonstrating that HCMTL's embedding-based connections do not significantly increase network size. Both HMTL and HCMTL have approximately three times more parameters than STL and SMTL, leading to three times longer training times. HCMTL achieves similar training times to HMTL because HCMTL uses observed target values from the training data to pass information between segments. As a result, later segments can directly retrieve relevant information from the data to initialize themselves. This allows HCMTL to train in the same manner as other models, using random shuffling in mini-batch gradient descent. 

Despite the differences in training time, all models exhibit similar inference times. This is because, during inference, we maintain the mini-batch size equal to the number of basins, ensuring that each mini-batch contains segments starting at the same time from all basins. This enables segment inference in chronological order across all basins. This approach benefits HCMTL, as HCMTL naturally requires inference to follow temporal order to pass information between segments. As a result, HCMTL achieves inference times comparable to other models.

\subsection{STL and SMTL with Larger Hidden States}
\begin{table}[htbp]
    \caption{STL and SMTL with Larger Hidden States}
    \label{tab:3xhiddens}
    \resizebox{\linewidth}{!}{
    \begin{tabular}{c|c|c|c|c}
        \multicolumn{5}{c}{USA, 7 Years, Window: 365, Stride: 182, hidden states: 768}\\
        \hline
        Model& Parameters& $\downarrow$ RMSE & $\uparrow$ NSE$_\text{mean}$ & $\uparrow$NSE$_\text{median}$\\
        \hline
        STL  &2492161 &1.254 &0.656&0.745\\
        SMTL &2493699 &1.240 &0.627 &0.761\\
        \hline
    \end{tabular}
    }
\end{table}
In the experiments, HCMTL uses three LSTM models, while STL and SMTL use only one. This raises the question of whether the comparison is fair. To explore this, we increased the hidden state size of STL and SMTL from 256 to 768 and trained six models for each to do ensemble learning. 

Table \ref{tab:3xhiddens} shows that increasing the hidden state size significantly raised the number of network parameters, with STL and SMTL reaching approximately 2,493,000 parameters—2.7 times more than HMTL and HCMTL, and 8.1 times more than STL and SMTL with 256 hidden state size (see Table \ref{tab:computational_complexity}). Despite this increase, STL and SMTL remained below the performance in RMSE and showed only minor improvements in NSE values (see Table \ref{tab:overall_results}). These findings indicate that simply increasing the hidden state size does not improve STL and SMTL's performance, underscoring the effectiveness of HCMTL's architecture.

\subsection{Results (RMSE)}
\subsubsection{Handling Short Time Series Data} \label{appendix:datasize}
\begin{figure}[htbp]
    \centering
    \includegraphics[width=\linewidth]{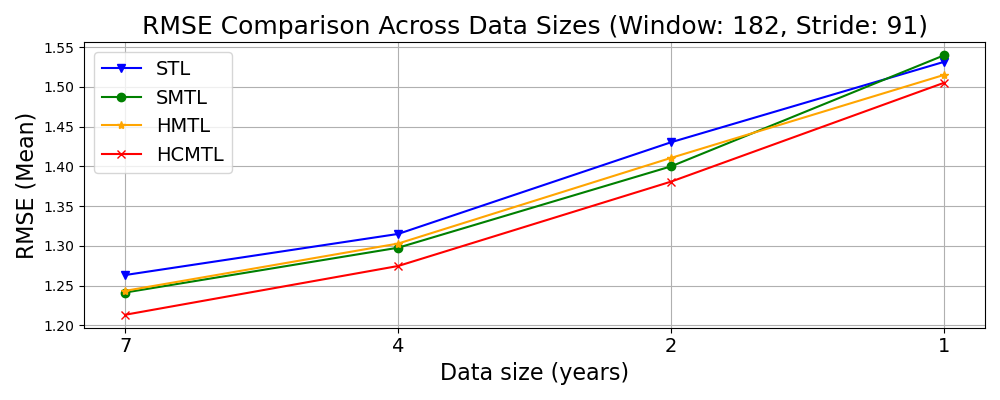}
    \caption{All models show higher RMSEs as training data size decrease.}
    \label{fig:datasize_usa_win182}
\end{figure}
In hydrology, some data-scarce basins may have very limited observed streamflow, sometimes less than one year. Without data augmentation, each basin would only produce one sample segment, making the model prone to overfitting. A common approach to address this challenge is to reduce the sliding window size and stride to generate more segments. In this experiment, we use a window size of 182 days with a stride of 91, generating three segments per basin from one year of training data. Figure \ref{fig:datasize_usa_win182} shows that HCMTL consistently outperforms other baselines, even with just one year of data. This demonstrates HCMTL's robustness in handling short time series data.

\subsection{Results (NSE)}
\subsubsection{Nash-Sutcliffe Efficiency (NSE)}
The Nash-Sutcliffe Efficiency (NSE) is a widely used metric in hydrology, comparing model performance to a mean-based baseline. NSE values range from 1 (indicating a perfect match) to negative values (indicating performance worse than the mean-based baseline that predicts the mean of the training data for the test data). In this section, we calculate the NSE for each basin to assess model performance on a basin-by-basin basis.
\begin{equation}
    \text{NSE} = 1 - \frac{\sum_{t=1}^{T} \left( y_t - \hat{y}_t \right)^2}{\sum_{t=1}^{T} \left( y_t - \bar{y} \right)^2}
    \label{eq:nse_appendix}
\end{equation}

\subsubsection{Basin-by-Basin NSEs}
\begin{figure}[htbp]
    \centering
    \includegraphics[width=\linewidth]{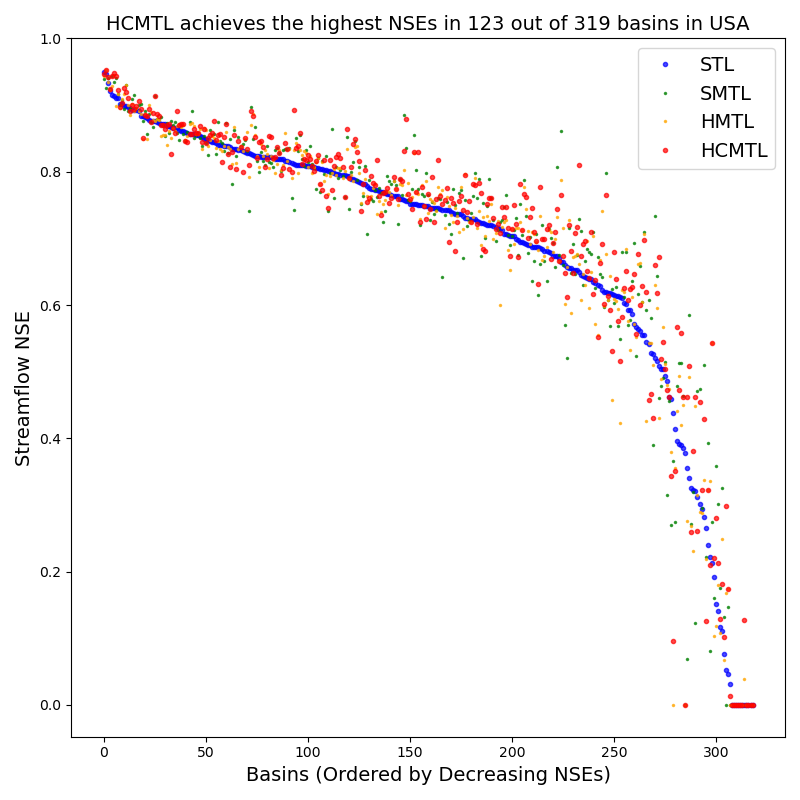}
    \caption{Basins are sorted by decreasing NSE values for the STL models, and the NSE values for both the baselines and the proposed HCMTL model are plotted for each basin.}
    \label{fig:basin_r2_trend}
\end{figure}
We sort the basins by decreasing NSE values for the STL models and then plot the NSEs for both the baselines and the proposed HCMTL model for each basin. Figure \ref{fig:basin_r2_trend} shows that HCMTL generally outperforms the baselines, achieving the highest NSEs in 123 out of 319 basins in the United States.

\subsubsection{Empirical Cumulative Distribution Function of the NSEs}
The Empirical Cumulative Distribution Function (ECDF) is a non-parametric estimate of the cumulative distribution function (CDF) for a dataset. It shows the proportion or percentage of data points that are less than or equal to a given value. The ECDF makes no assumptions about the underlying probability distribution of the data and accumulates probabilities as you move across the data range.

\begin{figure}[htbp]
    \centering
    \includegraphics[width=\linewidth]{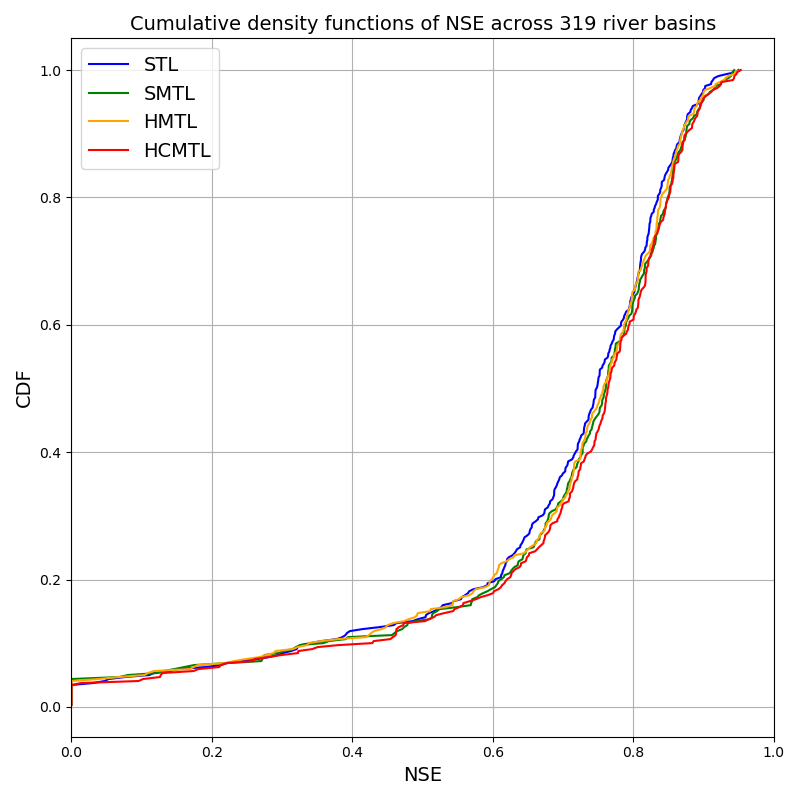} 
    \caption{Empirical Cumulative Distribution Functions (ECDF) of NSE values for 319 U.S. basins. The X-axis represents NSE values, while the Y-axis (ranging from 0 to 1) shows the proportion of basins with NSE values below a given threshold. The curve starts at (0, 0), meaning none of the data is below the smallest value, and ends at (1, 1), indicating all the data is below or equal to the maximum value. A steep slope suggests many basins have similar NSE values, while a gradual or flatter slope indicates more variability. For an NSE ECDF, a model with better performance will have its curve closer to 1.0 NSE, indicating a larger proportion of basins with higher NSE values.}
    \label{fig:basin_r2_cdf}
\end{figure}

Figure \ref{fig:basin_r2_cdf} presents the Empirical Cumulative Distribution Functions (ECDF) of NSE values for 319 U.S. basins. It shows that approximately 75\% of basins have NSE values ranging from 0.6 to 0.95, highlighting the effectiveness of both the baselines and the proposed HCMTL in streamflow modeling. Among the models, HCMTL's curve has a steeper slope and is closer to 1.0, indicating that HCMTL achieves the best performance, with a larger proportion of basins exhibiting higher NSE values.

\end{document}